%% file: main.tex
\documentclass[format=acmsmall, review=false, screen=true]{acmart}

\usepackage{booktabs} 
\usepackage{todonotes} 
\usepackage{tabularx}
\usepackage{multirow}
\usepackage[bottom]{footmisc}
\usepackage[para,online,flushleft]{threeparttable}

\newcommand{\neuraghe}{\textsc{NEURAghe}} 

\DeclareGraphicsExtensions{.pdf,.jpeg,.png}
\graphicspath{{./images/}}

\usepackage[ruled]{algorithm2e} 

\SetAlFnt{\small}
\SetAlCapFnt{\small}
\SetAlCapNameFnt{\small}
\SetAlCapHSkip{0pt}
\IncMargin{-\parindent}
\acmJournal{TRETS}

\setcopyright{acmlicensed}

\acmDOI{0000001.0000001}

\received{December 2017}

\begin{document}
\title[\neuraghe]{\neuraghe{}: Exploiting CPU-FPGA Synergies for Efficient and Flexible CNN Inference Acceleration on Zynq SoCs}

\author{Paolo Meloni}
\affiliation{
  \institution{Universit\`{a} di Cagliari}
  \city{Cagliari}
  \country{Italy}
}
\email{paolo.meloni@diee.unica.it}
\author{Alessandro Capotondi}
\affiliation{
  \institution{Universit\`{a} di Bologna}
  \city{Bologna}
  \country{Italy}
}
\email{alessandro.capotondi@unibo.it}
\author{Gianfranco Deriu}
\affiliation{
  \institution{Universit\`{a} di Cagliari}
  \city{Cagliari}
  \country{Italy}
}
\affiliation{
  \institution{T3LAB}
  \city{Bologna}
  \country{Italy}
}
\email{gianfranco.deriu@unica.it}
\author{Michele Brian}
\affiliation{
  \institution{T3LAB}
  \city{Bologna}
  \country{Italy}
}
\email{michele.brian@t3lab.it}
\author{Francesco Conti}
\affiliation{
  \institution{Universit\`{a} di Bologna}
  \city{Bologna}
  \country{Italy}
}
\email{f.conti@unibo.it}
\affiliation{
  \institution{ETH Zurich}
  \city{Zurich}
  \country{Switzerland}
}
\email{fconti@iis.ee.ethz.ch}
\author{Davide Rossi}
\affiliation{
  \institution{Universit\`{a} di Bologna}
  \city{Bologna}
  \country{Italy}
}
\email{davide.rossi@unibo.it}
\author{Luigi Raffo}
\affiliation{
  \institution{Universit\`{a} di Cagliari}
  \city{Cagliari}
  \country{Italy}
}
\email{raffo@unica.it}
\author{Luca Benini}
\affiliation{
  \institution{Universit\`{a} di Bologna}
  \city{Bologna}
  \country{Italy}
}
\email{luca.benini@unibo.it}
\affiliation{
  \institution{ETH Zurich}
  \city{Zurich}
  \country{Switzerland}
}
\email{lbenini@iis.ee.ethz.ch}

\begin{abstract}
Deep convolutional neural networks (CNNs) obtain outstanding results in tasks that require human-level understanding of data, like image or speech recognition.
However, their computational load is significant, motivating the development of CNN-specialized accelerators.
This work presents \neuraghe{}, a flexible and efficient hardware/software solution for the acceleration of CNNs on Zynq SoCs.
\neuraghe{} leverages the synergistic usage of Zynq ARM cores and of a powerful and flexible Convolution-Specific Processor deployed on the reconfigurable logic.
The Convolution-Specific Processor embeds both a convolution engine and a programmable soft core, releasing the ARM processors from most of the supervision duties and allowing the accelerator to be controlled by software at an ultra-fine granularity.
This methodology opens the way for cooperative heterogeneous computing: while the accelerator takes care of the bulk of the CNN workload, the ARM cores can seamlessly execute hard-to-accelerate parts of the computational graph, taking advantage of the NEON vector engines to further speed up computation.
Through the companion NeuDNN SW stack, \neuraghe{} supports end-to-end CNN-based classification with a peak performance of 169 Gops/s, and an energy efficiency of 17 Gops/W.
Thanks to our heterogeneous computing model, our platform improves upon the state-of-the-art, achieving a frame rate of 5.5 fps on the end-to-end execution of VGG-16, and 6.6 fps on ResNet-18.
\end{abstract}

%
%


%
%


\maketitle

\renewcommand{\shortauthors}{P. Meloni et al.}

\input{01_intro.tex}
\input{02_related.tex}

\input{03_architecture.tex}
\input{04_neudnn.tex}

\input{05_results.tex}
\input{06_conclusion.tex}

\bibliographystyle{ACM-Reference-Format}
\bibliography{bibliography}

\end{document}

%% file: 01_intro.tex
\section{Introduction}
\label{sec:intro}

In the last few years, Deep Convolutional Neural Networks have become the go-to solution for most tasks that require human-level understanding of data.
Thanks to their outstanding results, they represent the state-of-the-art in image recognition \cite{Krizhevsky2012,WuYSDS15,He2015}, face detection \cite{Taigman2014}, speech recognition \cite{HannunCCCDEPSSCN14}, text understanding \cite{Weston2014MemoryNetworks, Weston2016DialogbasedLanguageLearning} and artificial intelligence in games \cite{Mnih2015Humanlevelcontrol, Zastrow201621} among other tasks.
The big success of CNNs over the last few years can be attributed to the availability of large datasets and to the increasingly large amount of computational power available in General-Purpose Graphic Processing Units (GP-GPUs) to train these networks.

Training of CNNs has been traditionally performed on large servers of General Purpose Processors (GPP) or GP-GPUs since the large variety of algorithms and software frameworks coupled with their high computational complexity require the exploitation of general purpose processors. On the other hand, the regular computational structure of CNN inference, coupled with the inherent parallelism of the convolution operator dominating their computation time, has resulted in a large number of dedicated accelerators much more energy-efficient than general purpose processors \cite{TPU,NeuroStream,NVDLA}. Two notable example of such dedicated accelerators are the Google Tensor Processing Unit (TPU) \cite{TPU}, and the NVIDIA Deep Learning Accelerator (NVDLA) recently released open-source by NVIDIA. Originally designed for the inference task, and given the importance of the learning, Google announced a second, more flexible version supporting floating point operations, also suitable for training of CNNs \cite{TPUv2}. High-level tools allow to efficiently implement CNNs on these architectures starting form the CNN model's description created in training frameworks such as PyTorch, TensorFlow or Caffe, abstracting the complexity of the CNN models to the end user.

%

Embedded architectures for CNN acceleration mainly focus on the inference task, requiring a workload much smaller and regular than training algorithms, and much smaller dynamic and arithmetic precision (e.g. 16-bit fixed point).
A widely used category of embedded platforms for CNNs is that of systems-on-chip (SoC) integrating multi-core processors such as ARM Cortex A accelerated with embedded GP-GPUs such as NVIDIA Kepler \cite{TegraK1} or Maxwell \cite{TegraX1}, also featuring LPDDR memory interfaces to sustain the huge memory bandwidth typical of CNNs. Other systems rely on embedded heterogeneous SoCs built around ARM Cortex processors and FPGAs, such as the Xilinx Zynq \cite{Zynq}, Xilinx Ultrascale+ \cite{Ultrascale+}, and Altera Arria10 \cite{Arria10}. These architectures allow to integrate powerful and efficient accelerators on the reconfigurable logic, exploiting spatial computation typical of application specific integrated circuits (ASIC) rather than thread-level parallelism typical of GP-GPUs. Although high-level synthesis flows allow to implement annotated ANSI-C and OpenCL programs on these heterogeneous systems, plugs to the training environments have been announced by the main FPGA vendors but not made available to developers so far. Several dedicated accelerators have also been proposed in the embedded domain both from companies such as Movidius \cite{Movidius} and from the research community \cite{ShiDianNao,Origami,Eyeriss}, outperforming programmable solutions in both performance and energy efficiency. However, the deployment of these accelerators on real application environments has not been demonstrated, yet.

In this work we propose a CNN accelerator based on the Xilinx Zynq Z-7045 SoC. The proposed accelerator features an operating frequency of 140 MHz resulting into a performance up 169 GOPS and an energy efficiency up to 17 GOPS/W on end-to-end CNNs. A peculiar feature of the proposed accelerator relies on the presence of one controlling programmable soft-processor on the FPGA which manages the execution of complex CNNs on the Zynq SoC. This approach, which moves the intelligence closer to the compute engine implemented on the FPGA, enables an asynchronous execution model for the proposed accelerator, where the ARM Cortex A9 processor is released from any supervision duty after offloading the commands to the accelerator for the execution of the convolutional layer. This computational paradigm allows to implement a software pipeline where the highly optimized hardware accelerator executes the convolutional layers, while the ARM cores are responsible for the execution of fully-connected layers and data marshaling. Our approach fully leverages the synergy between the A9 cores and the FPGA, heavily exploiting the NEON vector engines to speed up the execution of the software layers, and achieving a very balanced execution time breakdown and very high utilization of all computing resources available on the SoC (hard vector engines and soft FPGA datapath). The accelerator comes with a software environment that allows to automatically generate the ARM host program and the correct memory layout of the weights trained with standard frameworks. The proposed hardware/software architecture is demonstrated through the deployment of the VGG-16 and ResNet-18 CNNs, trained using the Caffe training framework. The evaluated benchmarks achieve a frame rate of 5.5 FPS and 6.6 FPS on the proposed accelerator, respectively, which significantly improves performance and energy efficiency of end-to-end convolutional neural networks over the best-in-class CNN accelerators implemented on the Zynq z-7045 SoC reported in literature. The proposed approach is fully flexible and portable. On the one hand, it allows to easily implement any kind of CNN models fully exploiting the hardware and software capabilities of the Z-7045 SoC; on the other hand, it also eases the porting with big performance benefits to next-generation Ultrascale+ SoC.
These SoCs feature a bigger and faster FPGA on the programmable logic (PL), which would allow to host two convolutional engines running at 200 MHz, and they also feature a more powerful processing system (PS) based on a quad-core ARM Cortex A53 processor.

The rest of the paper is organized as follows. Section \ref{sec:related} presents an overview of the state of the art of CNN architectures based on FPGA. Section 3 provides an overview of the computational model of CNNs. Section 4 describes the architecture of the proposed CNN accelerator. Section 4 gives an overview of the software framework that generates the code for the SoC and organize the weights according to the layout required by the accelerator. Section 5 details the implementation of the two CNNs used as use-cases. Section 6 provides a quantitative comparison with the other recently published FPGA CNN accelerators.

%% file: 02_related.tex
\section{Related Work}
\label{sec:related}

Following the explosion of applications of deep learning algorithms based on CNNs, both academia and industry have focused a significant part of their efforts in the deployment of these algorithms on FPGAs. The hierarchical, relatively simple structure of CNNs, mainly composed of accumulated convolutions with a pre-trained set of filters make them highly suited for FPGA implementation, mainly due to two reasons. First, the large amount of digital signal processing blocks (DSP blocks) enables efficient implementation of the multiply and accumulate elements representing the core of the convolution kernels. Second, as opposed to software programmable solutions such as CPUs and GP-GPUs, the surrounding logic can be adapted to massively exploit the spatial parallelism typical of hardware accelerators, and to customize the local and global memory accesses optimizing them to match the desired computational model.

Several works have tackled the problem of efficiently mapping CNNs onto FPGAs in several application domains which include acceleration of mainstream processors in data-centers, high-end embedded systems running state of the art CNN models, and deeply embedded systems running simpler CNN models that exploit strong quantization of weights to improve performance and energy efficiency at the cost of retraining and classification accuracy. In this section we give an overview of the works that relates more closely with the proposed FPGA accelerator.

Zhang et. al. \cite{Caffeine} proposed Caffeine, a hardware/software library to efficiently accelerate CNNs on FPGAs. Caffeine leverages a uniformed convolutional matrix multiplication representation targeting both computation-intensive convolutional layers and communication-intensive fully connected layers of CNN which maximizes the underlying FPGA computing and bandwidth resource utilization. CNN implementations based on Caffeine are implemented with the Xilinx SDAccel high-level synthesis tool integrated in the Caffe learning framework. The implementation of two average-complexity CNN models such as VGG and AlexNet has been evaluated with Caffeine achieving a peak performance of 365 GOPS on Xilinx KU060 FPGA and 636 GOPS on Virtex7 690t FPGA.

Similarly, Ma et. al. \cite{RTLCompiler} presented an RTL-level CNN compiler that generates automatically customized FPGA hardware for the inference tasks of CNNs from software to FPGA. The approach proposed by \cite{RTLCompiler} relies on a template accelerator architecture described in Verilog including all the main functions employed by CNNs such as convolutions, pooling, etc, which are automatically customized at design time to match the requirements of the target CNN model. This approach allows to exploit the full benefits of low-level RTL design (i.e. frequency, area) while relying on flexible customization which starts from the output of the Caffe learning framework. The proposed methodology is demonstrated with end-to-end FPGA implementations of complex CNN models such as NiN, VGG-16, ResNet-50, and ResNet-152 on two standalone Intel FPGAs, Stratix V and Arria 10, providing average performance up to 720 GOPS.

While these two frameworks provide huge performance gains leveraging large FPGA devices such as Virtex7 and Arria 10 FPGAs, they mainly target server applications exploiting batching to improve memory access performance and bandwidth utilization. 
This approach is not suitable for the embedded applications where cheap and compact SoCs integrating embedded processors and FPGAs are desirable, and images have to be processed in real-time. In this embedded domain, most recent works exploit the capabilities of Xilinx Zynq Z-7045 SoC, integrating a dual-core Cortex A9 processor operating up to 800 MHz and reconfigurable logic featuring 900 DSP slices.

Venieris et. al. \cite{LatencyDriven} presented a latency-driven design methodology for mapping CNNs on FPGAs. As opposed to previous presented approaches mainly intended for bandwidth-driven applications, this work targets real-time applications where the batch size is constrained to one. The proposed design flow employs transformations over a synchronous dataflow modelling framework together with a latency-centric optimization procedure to efficiently explore the design space targeting low-latency designs. This methodology, which relies on Xilinx high-level synthesis tools for mapping (i.e. Vivado HLS) provides extremely high resource utilization (i.e. the totality of the DSP slices of the Xilinx Zynq Z-7045 are employed). However, it has been demonstrated on a relatively simple CNN such as AlexNet, and on a very regular one such as VGG16 featuring only 3x3 kernels, providing a peak performance of 123 GOPS. This suggests the current limitations of HLS tools with respect to the template-based approach based on programmable or customizable RTL accelerators proposed in other architectures \cite{RTLCompiler}\cite{Snowflake}\cite{GoingDeeper}, including the one proposed in this work.

SnowFlake \cite{Snowflake} exploits a hierarchical design composed of multiple compute clusters. Each cluster is composed of four vectorial compute units including a vectorial MAC, vectorial max, a maps buffer, weights buffers and trace decoders. SnowFlake provides a computational efficiency of 91\%, and an operating frequency of 250 MHz (best-in class for CNN accelerators on Xilinx Zynq Z-7045 SoC). However, although the vector processor-like nature of the accelerator is very flexible, delivering significant performance also for 1x1 kernels, it prevents to fully exploit of spatial computation typical of application specific accelerators, which leads to overheads due to load/store operations necessary to fetch weights and maps from the buffers. This is highlighted by the very poor utilization of the DSP slices available on the FPGA (i.e. only 256 over 900), and by the performance when executing end-to-end convolutional neural networks, which is lower than that of other architectures including the proposed one even though the operating frequency of the CNN engine is significantly higher.

Among CNN FPGA architectures, the precision of arithmetic operands plays a crucial role in energy efficiency. Although most of the architectures available in literature feature a precision of 16-bit (fixed-point)\cite{LatencyDriven,Snowflake,RTLCompiler} some reduced-precision implementations have been proposed recently, relying on 8-bit, 4-bit accuracy for both maps and  weights, exploiting the resiliency of CNNs to quantization and approximation \cite{GoingDeeper}. 

Qiu et. al. \cite{GoingDeeper} proposed a CNN accelerator implemented on a Xilinx Zynq platform exploiting specific hardware to support 8/4 bit dynamic precision quantization, at the cost of 0.4\% loss of classification accuracy. To improve the performance of fully connected layers, mainly limited by the off-chip bandwidth, the architecture employs Single Value Decomposition (SVD) to reduce the memory footprint of the weights. The design was evaluated on a VGG-16 network featuring SVD on the first fully connected layer, and achieves a performance of 187.8 GOP/s and 137.0 GOP/s for CONV layers and full CNN under 150 MHz frequency respectively achieving 4.4 Frames Per Second (FPS).

Most extreme approaches to quantization exploit ternary \cite{Ternary} or binary \cite{FINN} neural-networks accelerators for FPGA. This approach significantly improves the computational efficiency of FPGA Accelerators, allowing to achieve performance level as big as 8 TOPS \cite{Ternary}. These improvements are due to the 32-bit multipliers that can be replaced by simpler multiplexer and 2's complement operators, while bandwidth for loading weights can be reduced drastically, by 8 to 16 times if we compare with widely used 16-bit fixed point accelerators. The main issue related to binary and ternary accelerator is related to the training. While small networks like MNIST, CIFAR10, SVHN, GTSRB can reach good classification accuracy, the training is still a big challenge for larger networks such as VGG or ResNet \cite{Courbariaux2015a}.

In this work we target execution of state of the art CNNs leveraging 16-bit operands and weights hence not requiring retraining. Starting from the work proposed in \cite{Meloni}, we have improved flexibility introducing support for computing kernels different then convolutions. To this aim, we have integrated support for pooling and activation layers and we have implemented and tested tight interaction with the ARM-based processing system in the Zynq, as an instrument to implement end-to-end CNNs. 

The peculiarity of the proposed accelerator specifically lies in the execution model: as opposed to all previously published works based on the Z-7045 SoC, where the ARM processors are only responsible for controlling the execution of the CNN, our approach exploit interaction with the processing system (PS) in the Zynq, including the use of the powerful and flexible NEON accelerators, to execute fully connected layers of CNNs. Moreover, our approaches maps on the PS "irregular" computing patterns, that are hard to implement on hardware pipelines. \neuraghe{} also leverages an asynchronous offload mechanism to enqueue commands to the convolutional accelerators on the programmable logic (PL). This approach allows to implement a software pipeline which overlaps convolutional and fully connected layers fully exploiting the compute capabilities of the Z-7045 SoC significantly improving the performance over best-in-class CNN accelerators implemented on the Zynq z-7045 SoC reported in literature. The proposed approach is highly flexible and portable, and very promising when moving to next generation Zynq Ultrascale+ SoC where the PL is capable to host two convolutional engines operating at 200 MHz, and the PS is based on a more powerful quad-core ARM Cortex A53 processor.

%% file: 03_architecture.tex
\section{\neuraghe{} System Architecture}\label{sec:architecture}

\subsection{Target computational model}
\label{sec:comp_model}
Convolutional Neural Networks can generically be represented as directed graphs in which each edge represents a data tensor, and each node represents an operation (a \textit{layer}) transforming one or more inbound tensors into an outbound tensor.
Most often the data tensors considered in CNNs for image processing applications are three-dimensional, with one dimension representing different \textit{channels} or \textit{feature maps} plus two spatial dimensions; especially in the final layers of a CNN topology, some of these tensors can be ``collapsed'' to 1D vectors, where the spatial notion has been lost.
Operations performed in a node can range from convolutions, pooling, and fully connected layers (the most common ones), to generic operations such as tensor concatenation, to special-purpose ones in more exotic cases.
Convolutional layers transform a 3D tensor of size $N_\mathrm{\scriptscriptstyle i} \times h \times w$ into a new 3D tensor of size $N_\mathrm{\scriptscriptstyle o} \times h' \times w'$\footnote{In most CNN topologies, convolutional layers employ zero-padding of the input to make it so that $h'=h$ and $w'=w$. Convolutions can also use stride greater than 1 in one or both spatial  directions.} by means of a combination of convolutions operating on the spatial dimensions followed by a pointwise non-linear activation (often rectification).
The linear part of the layer is the following:
\begin{equation}
 \mathrm{for}\;k_\mathrm{\scriptscriptstyle o} \in 0\cdots N_\mathrm{\scriptscriptstyle o}-1,\quad
	\mathbf{y}({\scriptstyle k_\mathrm{\scriptscriptstyle o}}) =
	\mathbf{b}({\scriptstyle k_\mathrm{\scriptscriptstyle o}})
	+ 
	\sum_{k_\mathrm{\scriptscriptstyle i}=0}^{N_\mathrm{\scriptscriptstyle i}-1}
	\Big(\mathbf{W}({\scriptstyle k_\mathrm{\scriptscriptstyle o}}, {\scriptstyle k_\mathrm{\scriptscriptstyle i}}) * 
	\mathbf{x}({\scriptstyle k_\mathrm{\scriptscriptstyle i}})
	\Big)
	\label{eq:conv_layer}
\end{equation}
where $\mathbf{W}$ is the tensor of weights, $\mathbf{b}$ the one of biases, $\mathbf{x}$ is the tensor of input feature maps and $\mathbf{y}$ the one of output feature maps (before activation).
Fully connected layers have a similar structure, but they operate on 1D vectors (which can be flattened tensors) and the linear part of the layer is a full matrix-vector multiplication:
\begin{equation}
	\mathbf{y} =
	\mathbf{b}
	+ 
	\mathbf{W} \cdot 
	\mathbf{x}
	\label{eq:fc_layer}
\end{equation}

In most CNN topologies, convolutional layers (coupled with pooling) are responsible of the overwhelming majority of operations, and are typically compute-bound due to the high degree of data reuse offered by convolutions; fully connected layers, on the other hand, are responsible for much of the remaining operations, but they are memory-bound due to the absence of reuse.
To provide high throughput, a CNN accelerator must therefore be able to speed up the former layers and to hide as much as possible the cost of the latter, which are typically dominated by the memory traffic to fetch the weights.
Therefore we designed \neuraghe{} taking into account three primary objectives:
\begin{enumerate}
	\item support the deployment of arbitrary CNN topologies
    \item acceleration of critical compute-bound operations (i.e. convolutional layers)
    \item hiding of memory-bound operations (i.e. fully connected layers) by overlapping them with the compute-bound ones
\end{enumerate}
To meet these objectives, 
the \neuraghe{} platform employs a hybrid HW-SW scheme in which a \textit{general-purpose processor} (\textit{GPP}) cooperates with a \textit{convolution-specific processor} (\textit{CSP}).
The full CNN model is decomposed in the execution of each layer, which can take place either in the GPP or in the CSP, which is dedicated to accelerate the compute-bound convolution tasks and is able to execute also the operations that are more commonly coupled with convolution (activation, padding, pooling).

The CSP and GPP can work concurrently to maximize throughput; however, since most CNN topologies are predominantly sequential, it is sometimes difficult to overlap the execution of convolutional and fully connected layers pertaining to the same execution of the overall model, i.e. to the same input frame.
Luckily, in many common CNN topologies such as VGG, fully connected layers are only present at the end of the model.
This means that, in presence of a stream of input frames, it is often possible to overlap the execution of convolutional layers pertaining to frame $i+1$ with that of the final fully connected layers of frame $i$, effectively hiding the memory-bound operations.

\subsection{System architecture}
\begin{figure*}[ht]
\centering
\includegraphics[width=0.7\textwidth]{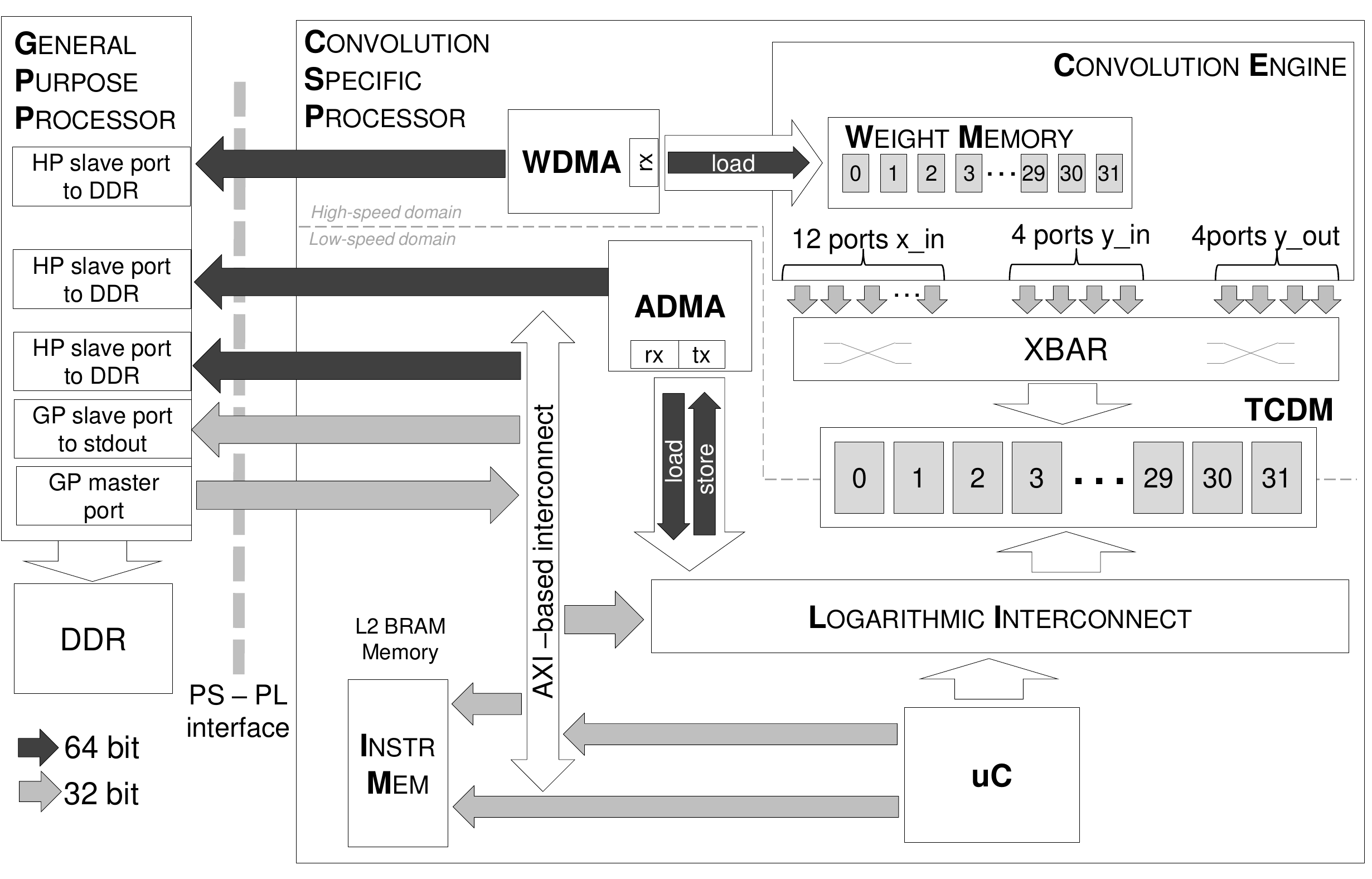}
\caption{Architectural template}
\label{architectural_template}
\end{figure*}

Figure \ref{architectural_template} reports the overall system-level organization of \neuraghe{}. 
It is built on top of a Xilinx Zynq SoC and it leverages both the dual Cortex-A9 processing system, which is used as general-purpose processor (\textit{GPP}), and the reconfigurable logic, which hosts the \textit{convolution-specific processor} (\textit{CSP}).
\neuraghe{} exploits two high-performance 64 bit ports for CSP-to-{GPP} communication (e.g. to access the memory-mapped off-chip DDR) and two general-purpose 32 bit ports for memory-mapped control of the \neuraghe{} architecture and standard output.
As detailed in Section \ref{sec:NeuDNN}, the {GPP} is used as an active partner in the heterogeneous computation of complex CNN topologies, carrying out tasks that would be accelerated less effectively on the programmable logic, such as memory-bound fully connected layers.


\subsection{Convolution-Specific Processor}
\label{sec:cluster}

The Convolution-Specific Processor is composed of several submodules, entirely described in synthesizable SystemVerilog HDL: a local tightly-coupled data memory (\textit{TCDM}) used to store activations and runtime data, a weight memory (\textit{WM}) a weight DMA controller to move weights to the {CSP} (\textit{WDMA}), an activation DMA to move activations in/out of the CSP (\textit{ADMA}), a simple microcontroller soft-core (\textit{$\mu$C}), and the inner nucleus of the {CSP}, the actual Convolution Engine (\textit{CE}) that embeds the sum-of-products units used to deploy convolutions on the reconfigurable logic.

The {CSP} architecture is centered around the local TCDM, which can be concurrently accessed by the uC, the ADMA, a slave port from the GPP, and the {CE}.
The {TCDM} is implemented with 32 banks of dual-port BRAM primitives, with one port dedicated to direct access from the CE by means of a simple crossbar (\textit{XBAR}), and the other shared between all the other master by means of a low-latency logarithmic interconnect \cite{igor} (\textit{LIC}), which arbitrates concurrent access from multiple masters to a single bank by granting only one request using a round-robin starvation free protocol.

The embedded microcontroller is based on a simple OpenRISC core (\cite{Gautschi}) coupled with an instruction memory that is accessible on the {GPP} memory map and is loaded at boot time with a resident runtime environment used to orchestrate the overall {CSP} operation, e.g. to offload jobs to the CE, program ADMA and WDMA data transfers, notify the GPP of the completion of a CSP job.
The resident runtime is thoroughly described in Section \ref{sec:NeuDNN}.

The CSP operates on two independent clock domains: the WM, the WDMA, the CE and the XBAR constitute a \textit{high-speed} domain, while the uC, the LIC and the ADMA operate in a \textit{low-speed} one.
The dual port banks of which the TCDM is composed are clocked with the two separate clocks according to the connection (high-speed for the CE ports, low-speed for the rest).
This allows to maximize throughput for the CE, while keeping full flexibility for the rest of the devices.

\subsection{Convolution Engine}\label{sec:archi_hwce}
The internal architecture of the CE is inspired from the design introduced by Conti~et~al.~\cite{conti_ultralowenergy_2015,IoT_Endpoint_SoC} as an accelerator of multi-core ultra-low-power system-on-chips.
The CE focuses on accelerating convolution-accumulation loops and uses the local TCDM as the source of input feature maps (\textit{x}) and the storage of output feature maps (\textit{y}), both fully and partially computed.

%
\begin{figure}[!h]
\centering
\includegraphics[width=0.8\columnwidth]{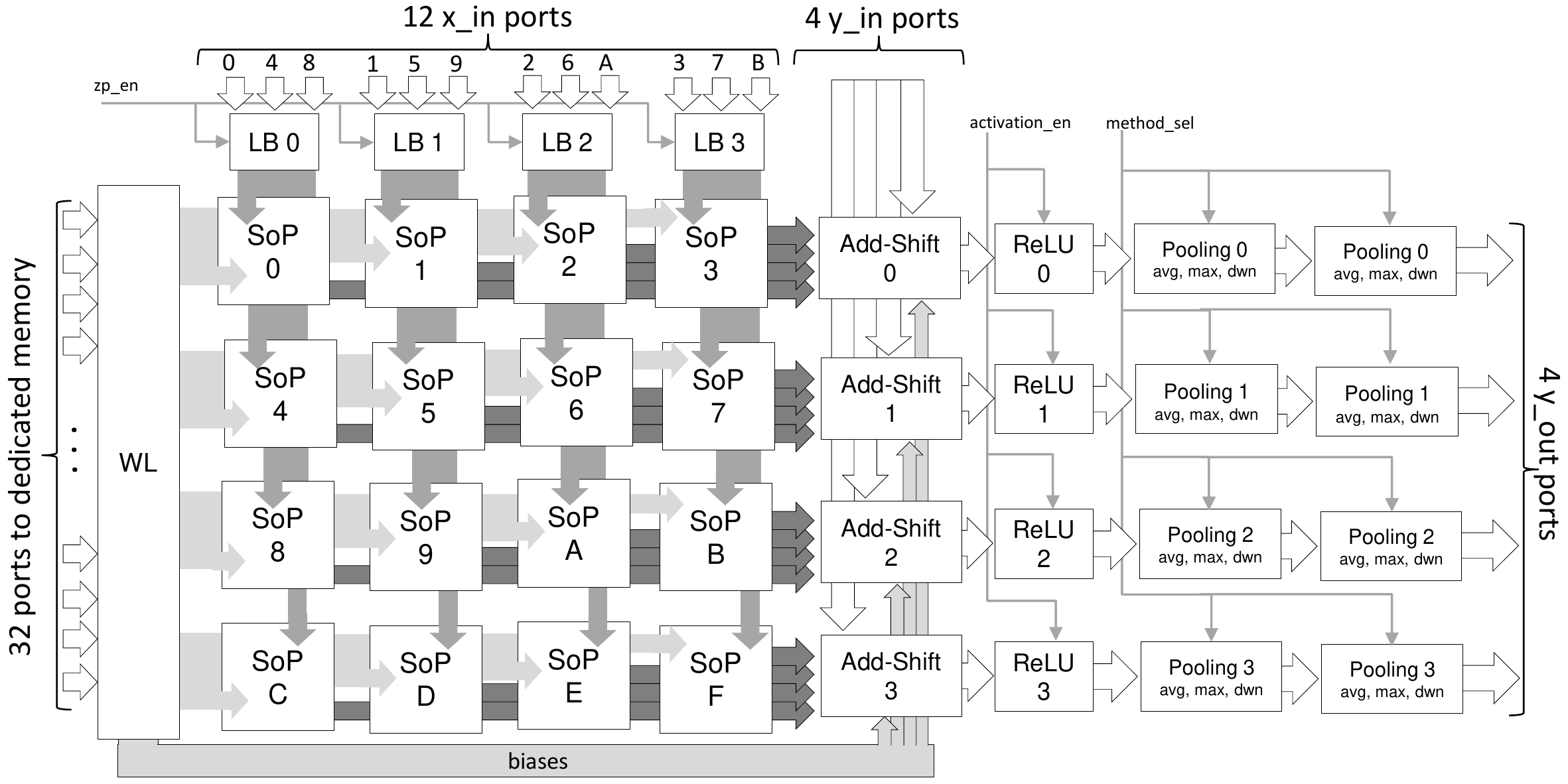}
\caption{CE organization}
\label{CE_architecture}
\end{figure}
As shown in Figure \ref{CE_architecture}, the CE features many connections to the TCDM:
\begin{itemize}
\item 12 \emph{x\_in} ports, that are used to read input features;
\item 4 \emph{y\_out} ports, that are used to write partial accumulation results or fully computed output features;
\item 4 \emph{y\_in} ports, that are used to read previous partial accumulation results. 
\end{itemize}
In each cycle of activity, the CE collects up to 12 input features through \emph{x\_in} ports 
and computes their contributions to 4 output features.
The input features \textit{x\_in} are loaded through a set of \emph{line buffers}, indicated with \emph{LB} in the diagram, which are used to cache the value of a few lines of the input image so that by loading a single new pixel per cycle an entire new window of the image can be dispatched to the Sum-of-Products (\textit{SoP}) modules to be convoluted with the weight filters.
In \neuraghe{}, the LB blocks are realized by means of shift registers.
As the CE works on 16-bit pixel data, each LB can be fed with two pixels per cycle obtained from the input port.
After an initial preloading phase, during which the first rows are filled, each LB produces two square convolution windows per cycle, centered on adjacent pixels.

The convolution windows are consumed by the SoP modules, which are the computational core of the accelerator. They apply the bi-dimensional filter kernel to the windows received by the LBs.
They are aggressively pipelined with a structure made up of trellises of multiply and add operations (a multiplier, an adder and two pipeline registers, see Section \ref{sec:sop_desc}) to maximize mapping efficiency on the FPGA DSP resources.
%
To cope with the throughput of two convolution windows per cycle produced by the LBs, each SoP module includes two sets of parallel trellises, for a total of 2$\times$N${}^2$ DSP blocks (where N is the size of the 2D kernel).

Pre-trained weights for a given kernel are loaded in a dedicated register file before the computation starts by a simple weight loader state machine (WL).
The WL is directly connected to the private weight memory, composed of a configurable number of BRAM banks and accessible in parallel to minimize weight loading overhead.
%
Referring to the scheme represented in Figure \ref{CE_architecture}, each row of the SoP matrix computes the contributions of input features to the same output feature. Thus, the outputs of the SoP modules in each row must be summed together. Moreover, since output values resulting from multiplication are wider than I/O connections, precision must be adapted to 16 bits with a shift operation, before connection to \emph{y\_out} ports. These operations are performed by the \emph{Adder-shifter} module, that is also in charge of the accumulation with previous partial results or with pre-trained bias values.


\subsection{Line buffers}
\begin{figure*}
\begin{center}%
\includegraphics[width=0.7\columnwidth]{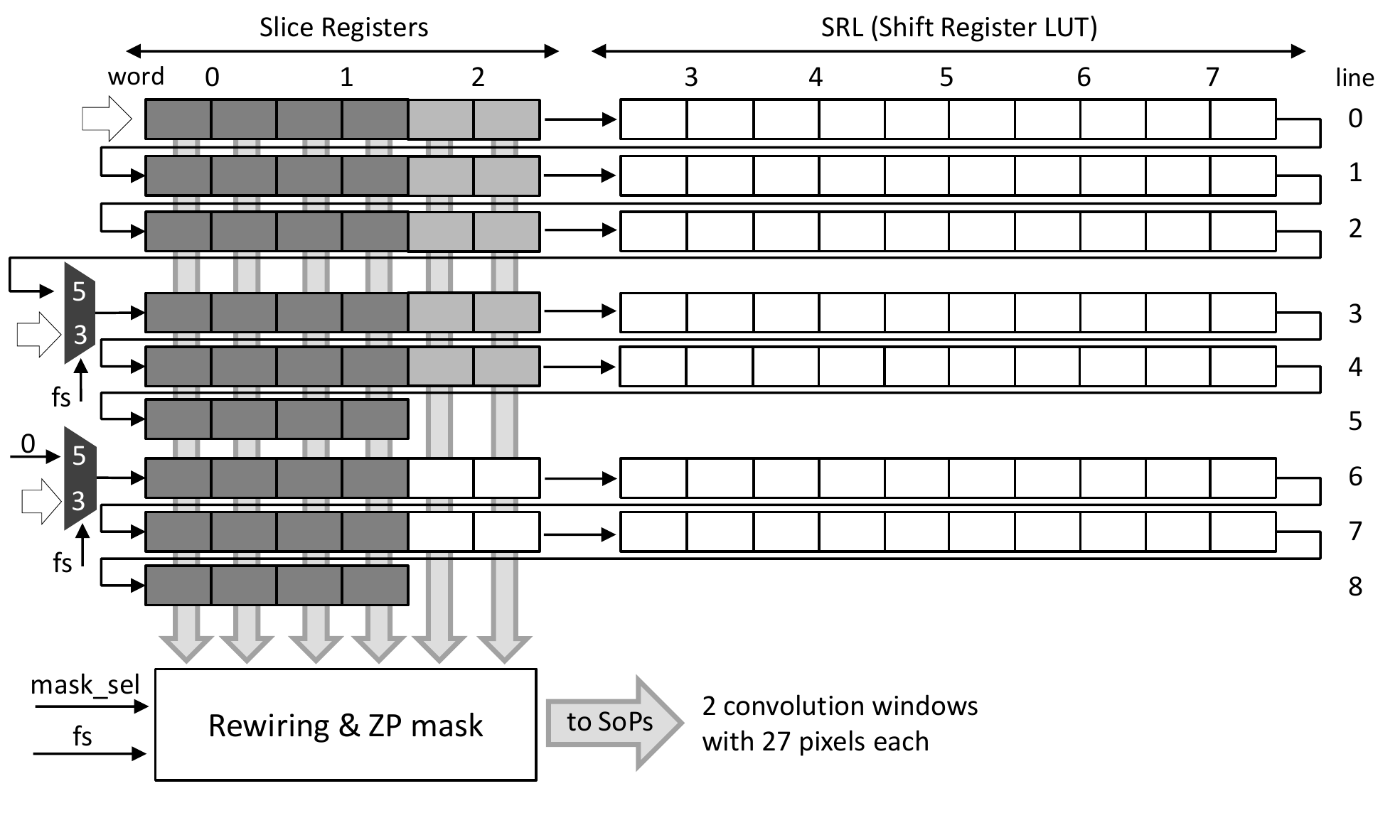}
\end{center}%
\caption{Reconfigurable line buffer architecture. Gray word slots are accessed only for 5x5 convolution windows, darker word slots are accessed for 3x3 convolution windows. Colored slots in lines from 0 to 4 are used for both configurations. In 5x5 configuration only one stream of input pixels is sent to the first line, while, in 3x3 configuration, the two muxes allow other two input streams to access line buffer from line 3 and 6. The first six words of each line are implemented with register slices, the others words are implemented with Xilinx SRL in order to save resources. Moreover, the content of colored locations are sent to a modules that performs a rewiring to connect slots to the right SoP and to apply zero-padding.}%
\label{fig:LB_subfigure}
\end{figure*}

In most CNNs, the size of the filtering kernel to be applied may be different for all the convolutional layers.
In order to improve the flexibility of our approach, first we have enriched the architecture, integrating line buffers that support different kernel sizes. 
The configuration proposed in Figure \ref{CE_architecture}, for example, can be reconfigured, by changing the behavior of line buffer modules (please see Fig \ref{fig:LB_subfigure}), at runtime by the processing elements in the cluster, to efficiently perform convolutions with 3$\times$3 or 5$\times$5 filters. \par
In the presented configuration, each SoP modules embeds 27 multipliers for each output pixel (54 in total, since SoP modules produce two output pixels per cycle). The 27 multipliers can be arbitrarily used, depending on the features of the convolution layer to be tackled, to perform either 3 different 3$\times$3 convolutions or one single 5$\times$5 convolution (leaving two multipliers unused in this case).\par
Moreover, to support reconfigurability, the line buffers are capable of switching at runtime between two operating modes, respectively reading one input stream (to be processed with 5$\times$5 filters) or three input streams (to feed the three 3$\times$3 filters computed by each SoP).
To this aim, the line buffer is equipped with an additional selection mechanism, controlled via software by means of memory-mapped registers accessible by the cores in the cluster, that can be reconfigured to set the line buffer functionality to the needed operating mode. 
In the first mode, the line buffer acquires one single stream of pixels and produces in output two windows of 25 pixels each, to be sent to the SoP modules. In the second mode, the shift register is partitioned in three independent regions, used by the line buffer to load three different streams corresponding to three different input features.\par
In Figure \ref{fig:LB_subfigure}, we show the line buffer internal structure, that allows the two mentioned operating modes. As may be noticed, some multiplexers are needed to re-configure the shifting path along the registers in the buffer. Moreover, some rewiring  circuitry is needed to select which pixels are part of a convolution window in the considered operation mode and must be forwarded to SoP modules. The buffer locations that correspond to convolution windows in the two modes are highlighted with different colors in the diagram. The same rewiring logic is used to implement zero padding on the input features before convolution, when needed. 

The re-configuration of the line buffer takes only one or two cycles and has to be performed at the beginning of the first CE activation in a convolution layer, thus it does not impact on performance. 

\input{03_02_SoP}
\input{03_01_pooling_relu}

%% file: 03_02_SoP.tex
\subsection{SoP modules}\label{sec:sop_desc}

\begin{figure}[t]
\begin{minipage}[t]{0.42\linewidth}
		\centering
		\vspace{-3.3cm}
    \includegraphics[width=\linewidth]{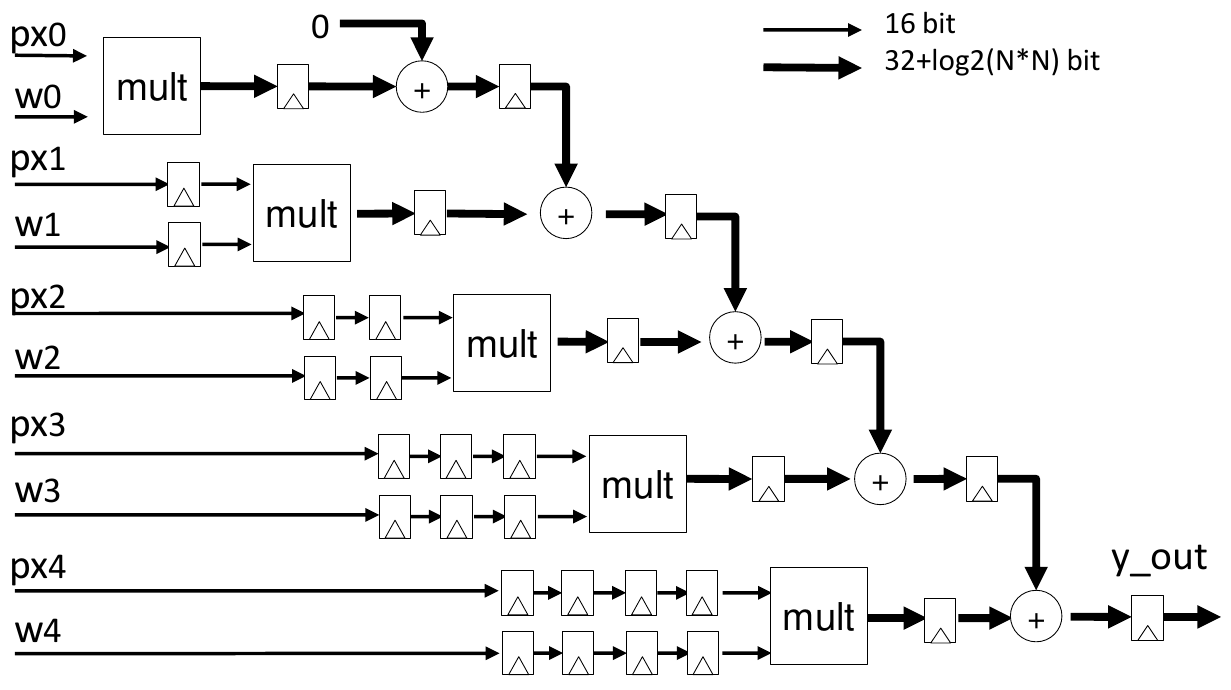}
    \caption{Single-trellis Sum-of-Products cascade.}
    \label{fig:SoP}
\end{minipage}
\hfill%
\begin{minipage}[t]{0.48\linewidth}
    \centering
		\includegraphics[width=\linewidth]{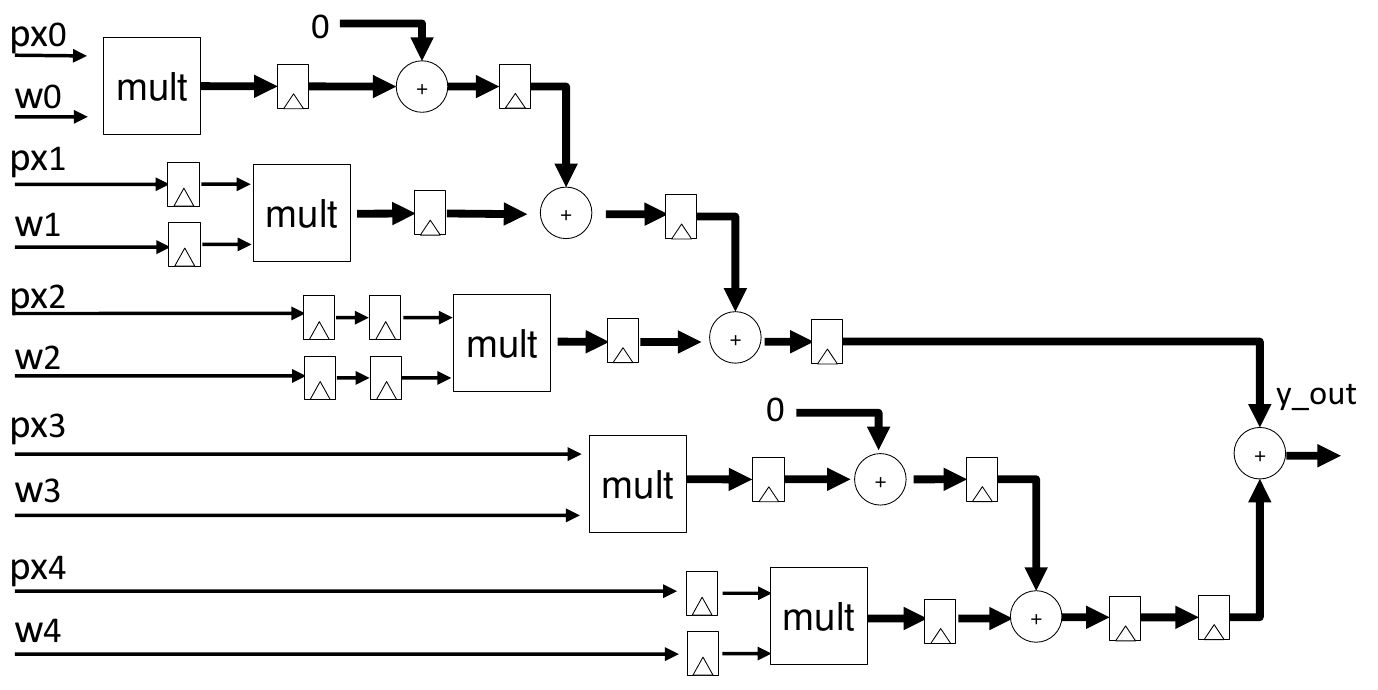}
		\caption{Multi-trellis Sum-of-Products cascade (with 2 trellises).}
		\label{fig:SoPasym}
\end{minipage}
\end{figure}

SoP modules are implemented using DSP48E1 primitives in the reconfigurable logic of the Zynq device.
The optimal implementation from the point of view of resource utilization would be a single trellis implemented as a cascade of DSP48E1 primitives, that exploits internal multipliers and adders to perform a multiply-and-accumulate operation and the input registers to keep the critical path independent from the size of the considered filtering kernel, as represented in Figure \ref{fig:SoP}.
However, in practice this single-trellis SoP couples many DSP48E1 resources tightly together, effectively imposing a restrictive placement constraint in the FPGA place \& route phase \footnote{DSP48E1 are placed in regular columns in the FPGA fabric in Xilinx Series-7 devices.}.
This can lead to a reduction of the maximum frequency or too long convergence time in a fairly congested design, in which the target is to use as many DSP48E1 blocks as possible.

To cope with this issue, the SoP structure can also be configured at design time to be partitioned in multi-trellis structures, whose outputs are summed together using a dedicated adder, as shown in Figure \ref{fig:SoPasym}.
Reducing the size of each trellis structure allows for more freedom when selecting the optimal mapping and placement of the resources, improving the overall implementation results and convergence time.
In the final \neuraghe{} design, we used a multi-trellis cascade with 6 trellises.

%% file: 03_01_pooling_relu.tex
\subsection{Pooling and ReLU module}
The CE architecture is also endowed with circuitry implementing computation kernels that may need to be executed right on the output of a convolutional layer in a CNN. 
Such hardware is placed at the output ports of the CE, as shown in Figure \ref{CE_architecture}, and can be controlled by the host processor using a set of dedicated memory mapped programmable registers.
First, the output pixels produced by each port of the convolution engine are streamed into a ReLU (Rectifier Linear Unit) block, that, when enabled, performs rectifier activation function on each pixel. 
Second, we have integrated on the accelerator a \emph{pooling} layer, that operates on the output streams produced by the convolution engine.
This layer is implemented by means of a shift register, that temporarily stores output pixels and compares values of pixels in square pooling windows. 
After comparison, according to the selected operating mode, the pooling layer outputs one single pixel per window.
The pooling layer can be set to perform \emph{max} pooling, \emph{average} pooling or a simple downsampling (statically selecting the pixel in a specific position in the window).
The default configuration of the pooling layer implements pooling over 2x2 windows. Two layers can be cascaded to implement 4x4 windows, alternatively activating at runtime only one or both layers, to dynamically switch between pooling schemes. Different configurations of the module, implementing different basic window sizes, can be chosen at design time.

%% file: 04_neudnn.tex
\section{NeuDNN: \neuraghe{} Deep Neural Network Software Stack}
\label{sec:NeuDNN}
The research field related with neural networks and deep learning represents a hot topic and it is freneticly growing.
New layers, new ML tools, and new neural networks topologies are released every day.
To tackle this fluid scenario it is crucial to provide a flexible and extensible programming interface that enables the reuse of existing hardware, software and algorithms.

\begin{figure}[t]
\centering
\includegraphics[width=0.85\columnwidth]{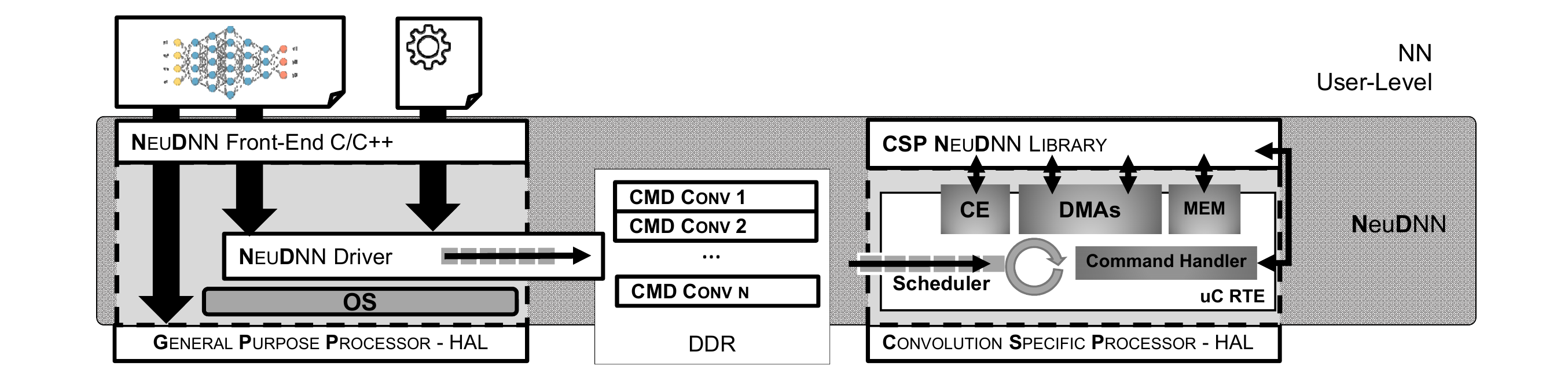}%
\caption{NeuDNN software stack overview.}%
\label{fig:neudnn}%
\end{figure}

To achieve these objectives we propose a complete and hardware-agnostic software stack, to enable an efficient implementation of Convolutional Neural Networks: the \neuraghe{} Deep Neural Network software stack (NeuDNN).
NeuDNN is an open-source\footnote{NeuDNN v1.0 will be publically released Q1 2018.} multi-target structured software stack which enables the user to write develop and reuse CNNs to be executed on the presented heterogeneous processing platform. NeuDNN sits on top Linux OS, thus the user is enabled to easily integrate in NN application 3rd Party and legacy software, like JPEG, and OpenCV libs.
Figure \ref{fig:neudnn} presents an overview of the whole software stack. 
It exploits the runtime design proposed Capotondi et al \cite{capotondi2017runtime} for hereterogenous many-core accelerator and provides a specialized implementation for FPGA-based accelerator.

NeuDNN consists of a C/C++ front-end, which can be used to specify and program CNN at software level, and of a back-end, that maps processing kernels to the hardware accelerator and controls their execution.
The back-end -- transparent to the user -- is composed of a NeuDNN Driver, used to offload computational task to the FPGA-accelerator, and of a Convolution Specific Processor resident RTE, executed by the $\mu$C, that receives requests from the driver and schedules autonomously the computation kernels on the Convolutional Engine and data transfers on the DMAs.

To implement a CNN, a user must develop a C/C++ code, exploiting NeuDNN APIs, and must define a simple configuration file describing the target computing platform (for example ARM SoC, or \neuraghe{}).
To load the data needed for the inference, weights and bias values, the user, helped by some migration tools provided by the NeuDNN, can easily import trained models from common ML tools like Tensorflow and Caffe.

\subsection{NeuDNN front-end}
The NeuDNN Front-End is a configurable C/C++ library for CNN deployment. It gives access to a set 
of statically linkable functions implementing pre-optimized layers and utilities for CNN development with no dependency from third party libraries.
The NeuDNN targets efficiently ARM class A processors and the \neuraghe{} architecture, supporting different activation format data types, such as 32-bit IEEE floating point and 16-bit fixed point.
Table \ref{tab:optKernels} lists the main computational kernels available as linkable C/C++ API.
By default, the library offers optimized implementations for all kernels and the data types deployable to the Generic Purpose Processor (GPP - in this particular case ARM class A cores).
All layers are optimized using OpenMP parallel programming model, to exploit parallelisms on the host-side, and ARM NEON vectorization, to exploit SIMD acceleration.
When Convolution Specific Processor (CSP) is available, some of those layers can be offloaded to the \neuraghe{} Convolutional Engine.
The CSP-based and the GPP-based implementations share the same APIs, thus the library may forward transparently the execution of the layer to most efficient engine.
To enable cooperative computation between the host and the CSP, the hardware accelerated \textit{Convolution*} layers support blocking and non-blocking semantics.
Like software \textsl{tasks}, multiple \textit{Convolution*} layers can be enqueued to the accelerator, while the host processor can be used to compute in parallel other layers.
These features are enabled by the lower level of NeuDNN software stack.
\footnotetext{Bigger convolutional filters can be implemented using a composition of these dimentions (see Section \ref{sec:results})}
\begin{table}[h]
\footnotesize
\renewcommand{\arraystretch}{0.7}
\begin{tabular}{l c c c c c l}
\textbf{Kernel} & \textbf{Dimensions} &\textbf{Stride} &\textbf{Data Type}	&\textbf{Deploy} &\textbf{Opt.} &\textbf{Note}\\
\toprule
\textit{Convolution} &Arbitrary &Arbitrary &float, 16-bit fixed &GPP&OpenMP, NEON &\\
\textit{Convolution*} &1$\times$1, 3$\times$3, 5$\times$5\footnotemark &4,2,1 &16-bit fixed &CSP&CE&Async, sync\\
\textit{Max Pooling} &Arbitrary &Arbitrary &float, 16-bit fixed &GPP&OpenMP, NEON &\\
\textit{Max Pooling*} &2$\times$2,4$\times$4&4,2,1 &16-bit fixed &CSP &CE &After \textit{Convolution*}\\
\textit{Avg Pooling} &Arbitrary &Arbitrary &float, 16-bit fixed &GPP&OpenMP, NEON &\\
\textit{Avg Pooling*} &2$\times$2,4$\times$4&4,2,1 &16-bit fixed &CSP &CE &After \textit{Convolution*}\\
\textit{Fully-Connected} &Arbitrary &Arbitrary &float, 16-bit fixed &GPP&OpenMP, NEON &\\
\textit{Add} &Arbitrary &Arbitrary &float, 16-bit fixed &GPP&OpenMP, NEON &\\
\textit{ReLU} &Arbitrary &---&float, 16-bit fixed &GPP&OpenMP, NEON &\\
\textit{ReLU*} &Arbitrary&--- &16-bit fixed &CSP&CE&After \textit{Convolution*}\\
\textit{Identity} &Arbitrary &Arbitrary &float, 16-bit fixed &GPP&OpenMP, NEON &\\
\textit{LRN} &Arbitrary &Arbitrary &float, 16-bit fixed &GPP&OpenMP, NEON &\\
\toprule
\end{tabular}
\caption{Optimized pre-defined NeuDNN Kernels}
\label{tab:optKernels}
\end{table}
\subsection{NeuDNN Back-End}
The NeuDNN back-end is distributed among the GPP and CSP.
The GPP side of the back-end is implemented as a driver, in charge of requesting the execution of APIs on the hardware accelerator and of the management of activation/data buffers. The driver takes care of the buffer marshaling and of the general transfers between the host DDR partition and the \neuraghe{} Convolution Specific Processor.
Actions on the accelerator are triggered by the driver by means of dedicated \textit{commands},
consisting in a set of meta-data structures that carry the information needed for the execution of the API (such as  weight array pointers, activation array pointers, etc.). Commands are stored in a shared FIFO queue mapped on the DDR address space.
Being NeuDNN implemented on top of the Linux OS, the DDR must be split in two partitions: one used by the OS as main virtual memory; and other one, unmapped and accessed by \texttt{/dev/mem}, contiguous and not paged, used to share data buffers between GPP and CSP.

The CSP-side is fully managed by a resident runtime, executed by the $\mu$C in the CSP, which is loaded and activated at the startup of the system, just after the load of the bitstream on the programmable logic.
The runtime, written in C, has direct access to the CSP HAL and is in charge of orchestrating data transfers from/to the local Convolutional Engine TCDM and triggers of CE activations. The runtime decomposes commands received by the GPP driver, requesting CNN basic operations such as Convolutions, Max Pool layers and ReLUs, into a scheduled track of elementary operations on the CE and on the two DMAs.
The used scheduling strategy is aggressively optimized to improve efficiency under limited bandwidth availability, using double-buffering and sliding window techniques to optimize the overlapping of computation with data transfers.

%% file: 05_results.tex
\section{Experimental Results}
\label{sec:results}
To evaluate the performance of \neuraghe{} and the flexibility of NeuDNN on real-world CNN topologies, we used our framework to implement two of the most commonly used ones: VGG-16 \cite{simonyan2014very} and ResNet-18 \cite{resnet-18}.
These two networks enable to show different computational approaches that can be supported using our framework, like computational pipelining and cooperative computation between the General Purpose Processor and the Convolution Specific Processor.
The results show up to 225 GOps/s\footnotemark delivered by the Convolution Specific Processor, and an end-to-end classification frame-rate on ImageNet up to 6.6 fps on ResNet-18, and 5.5 fps on VGG-16.
\footnotetext{
	As is often the case for CNNs, we count a multiply-accumulate as two operations. Using the notation of Section \ref{sec:comp_model} ($K_{h,w}$ denote the height and width of 2D filters), the performance of a convolutional layer is given by 
\[
	\mathrm{Perf} [\mathrm{GOps/s}] = {2 \times N_i \times N_o \times h' \times w' \times K_h \times K_w}\,\big/\,{t_\mathrm{execution}}
\]
}


As discussed in Section \ref{sec:architecture}, \neuraghe{} is deployed on a Xilinx Zynq Z-7045 SoC.
The two ARM Cortex A9 in the GPP are clocked at 800MHz, while  the CSP operates at 70MHz in the low-speed domain and at 140MHz in the high-speed one, including the CE. 
In this configuration, the GPP OS uses 744MB of the Xilinx PS DDR3, while the rest of the DDR3 (256MB) is used as contiguous shared memory accessible by both the GPP and the CSP.
The GPP is equipped with a Ubuntu 16.06 LTS OS (Linux Kernel 3.8) and the toolchain used for compilation was GNU GCC v5.4.

\input{05_01_hardware_implementation}

\subsection{VGG-16}

\begin{figure}[t]
\centering
\includegraphics[width=.75
\columnwidth]{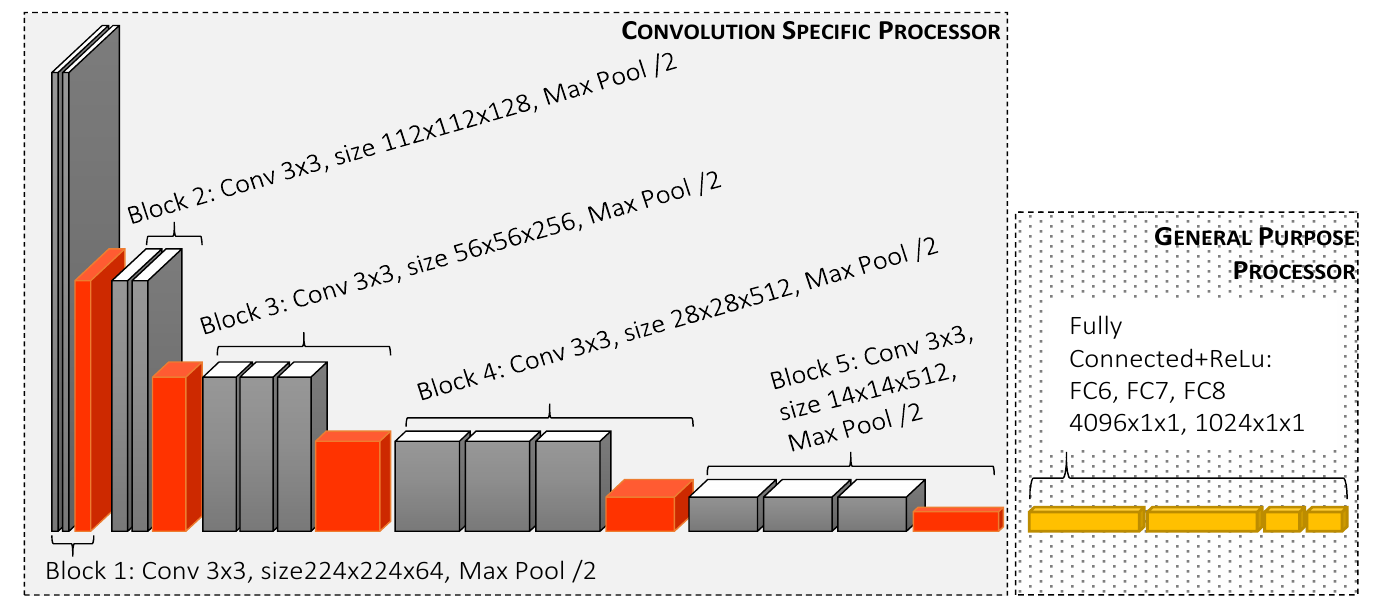}
\caption{VGG16 topology\cite{simonyan2014very}. Gray layers indicate 3$\times$3 convolutional layers; the Max Pooling reduction layers are in red, while the Fully Connected layers are in yellow. The two surrounding boxes highlight how the layers are mapped on the \neuraghe{} platform.}
\label{fig:vgg16}
\end{figure}

VGG is a deep convolutional neural network proposed by K. Simonyan \& A. Zisserman \cite{simonyan2014very}.
The model achieves up to 92.7\% top-5 test accuracy in ImageNet classification \cite{image-net}.
Figure \ref{fig:vgg16} shows the structure of VGG-16.
It consists of five computational \textit{blocks} followed by three fully-connected layers.
Each computational \textit{block} is composed of two or three 3$\times$3 convolutional layers followed by a max pooling reduction.

Compared to the standard VGG-16 network proposed by K. Simonyan \& A. Zisserman, in this work we exploited the SVD compression methodology proposed by Girschik~et~al.~\cite{GoingDeeper,girshickICCV15fastrcnn} for the first fully-connected layer (FC6).
This compression enables to reduce the memory footprint and the computational complexity of the FC6 layer of 3$\times$, with an accuracy loss smaller than 0.05\%.
\paragraph{VGG-16 \neuraghe{} deployment}
Mapping VGG-16 on \neuraghe{} is straightforward.
The five computational blocks can be enqueued to the CSP without any interaction with the GPP, while the fully connected layers can be fully executed on the GPP.
%
Compared to the original model, the \neuraghe{} implementation of VGG-16 requires two additional layers to manage the \textit{data marshaling} from/to the CSP - the first such operation is performed before the first VGG-16 block and the second between the last computational block and the first fully-connected layer.
The \textit{data marshaling} - as discussed in section \ref{sec:NeuDNN} - consists in the transfer of data from/to the OS-managed DDR section and the shared contiguous memory DDR partition, and the inter/deinter-lacing of activations.
The VGG-16 implementation uses 16-bit fixed-point data quantization for activations, weights, and bias, using Q5.11 format.

Table \ref{tab:vgg16-res} resumes activation size, measured execution time, and performance in GOps/s for all VGG-16 components (with the exception of data marshaling layer), divided in the respective computational blocks.
From the profiling, we can first observe that the total data marshaling overhead is below 13ms, i.e.  less than 5\% of the whole latency.
Together, all the VGG-16 computational blocks take  181ms, providing an average throughput of 169.7 GOps/s.
With the exception of the first convolutional kernel -- which offers a limited number of input features and then a limited possibility of parallelism for the Convolutional Engine -- the other convolutional kernels generate more than 100 GOps/s, with a peak performance of 225 Gops/s.
The fully-connected layers require on the 70 ms, with an average performance of 1.02 GOps/s. As we previously discussed, these layers are strongly dominated by the memory bandwidth.
The overall measured latency is 263.61 ms with a global average performance of 122.58 GOps/s.
\begin{table}[t]
\footnotesize
\renewcommand{\arraystretch}{0.7}
\begin{tabular}{l c c c }
&\textbf{Size}	 &\textbf{Time (ms)}	&\textbf{GOps/s}\\
\toprule
\textit{Marshaling}	&3$\times$224$\times$224&9.804&---\\
\toprule
\multirow{2}{*}{\textit{Block 1}} &64$\times$224$\times$224 &13.999   &12.38\\
&64$\times$224$\times$224 &32.784	   &112.8\\
\multirow{2}{*}{\textit{Block 2}} &128$\times$112$\times$112 &10.417	 &177.54  \\
&128$\times$112$\times$112 &18.14	 &203.92\\
\multirow{3}{*}{\textit{Block 3}} &256$\times$56$\times$56 &8.901	   &207.76\\
&256$\times$56$\times$56 &17.132	 &215.9\\
&256$\times$56$\times$56 &16.833	 &219.76\\
\multirow{3}{*}{\textit{Block 4}} &512$\times$28$\times$28 &8.68	&213.04 \\
&512$\times$28$\times$28 &16.578	   &223.12\\
&512$\times$28$\times$28 &16.428	 &225.12\\
\multirow{3}{*}{\textit{Block 5}} &512$\times$14$\times$14 &7.093	     &130.36\\
&512$\times$14$\times$14 &7.092	   &130.36\\
&512$\times$14$\times$14 &7.128	   &129.68\\
\toprule
\textit{Marshaling}	&512$\times$7$\times$7&2.9&---\\
\textit{FC 6-SVD}	&4096$\times$1$\times$1&32.554&0.917\\
\textit{FC 7}		&4096$\times$1$\times$1&29.46&1.138\\
\textit{FC 8}	  &1000$\times$1$\times$1&7.688&1.065\\
\toprule
\toprule
\textbf{Latency}	&&\textbf{263.61}&\textbf{122.58 (169.34\footnotemark)}\\
\toprule
\textbf{Pipelined}	&&\textbf{181.205}&\textbf{169.74}\\
\end{tabular}
\caption{VGG-16 measured performance on \neuraghe{}}
\label{tab:vgg16-res}
\end{table}
\footnotetext{Average GOps/s on convolutional layers.}

Thanks to the high flexibility of our proposed architecture and software stack, different execution models can be implemented to extract better performance.
Considering the common scenario where the input images are frames from a video stream, we can take advantage of the strong segregation of layers between the Convolution Specific Processor and the General Purpose Processor to improve the overall throughput of the VGG-16 applying a three-stage pipeline.
%
%
This is implemented by enqueuing the execution of convolutional blocks in asynchronous fashion, and letting the GPP execute the fully connected layers for frame $i-1$, while the convolutional blocks of frame $i$ are being computed by the CSP.
A third stage is added to remove the overhead of the first data marshaling from the critical path of the SW pipeline.
\paragraph{VGG-16 performance analysis}
\begin{figure}[t]
\begin{minipage}[t]{0.48\linewidth}
		\centering
    \includegraphics[width=0.8\linewidth]{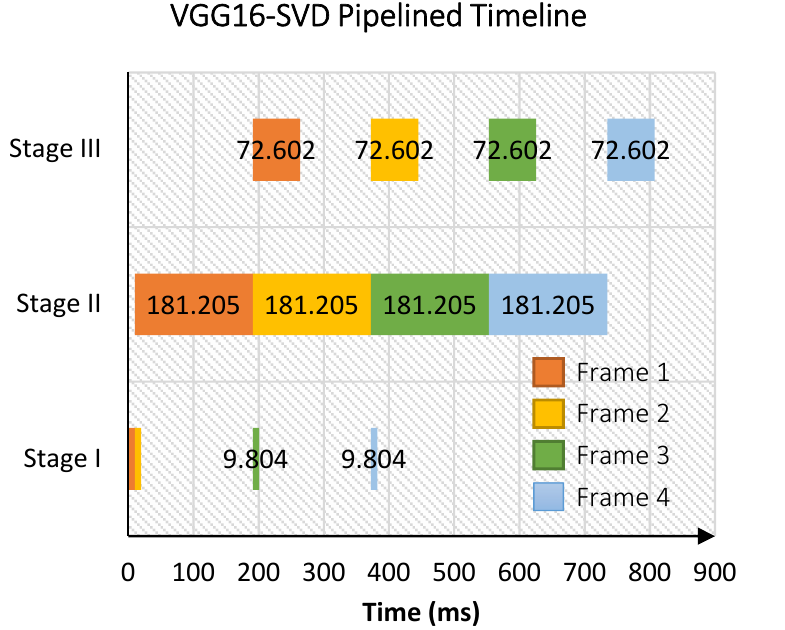}
    \caption{VGG16 four frame execution timeline}
    \label{fig:vgg16:pipelined}
\end{minipage}
\begin{minipage}[t]{0.48\linewidth}
    \centering
		\includegraphics[width=0.8\linewidth]{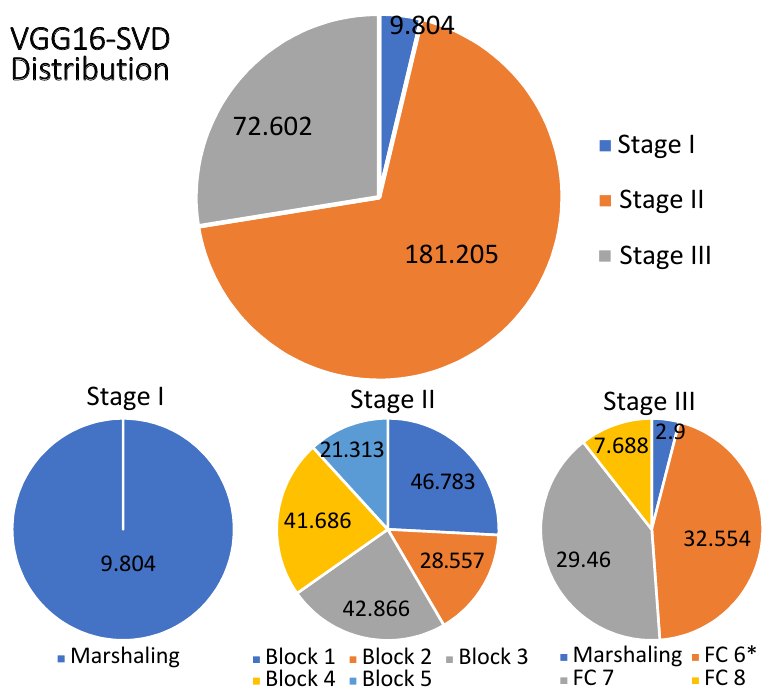}
		\caption{VGG16 pipeline time distribution in ms}
		\label{fig:vgg16:pipe-distr}
\end{minipage}
\end{figure}

The VGG16 is then split in three stages as follow:
\begin{itemize}
	\item \textbf{Stage I:} consists only of the the first data marshaling layer.
	\item \textbf{Stage II:} consists of all the computational blocks executed on the Convolution Specific Processor.
	\item \textbf{Stage III:} consists of all the rest of layers (marshaling, and fully-connected) executed on the General Purpose Processor.
\end{itemize}
A clear view of the execution in pipeline of VGG-16 is given by the Figure \ref{fig:vgg16:pipelined}.
The figure shows a real timeline, profiled on a \neuraghe{} board, of the execution of VGG-16 on 4 frames.
Figure \ref{fig:vgg16:pipe-distr} shows how the execution time are distributed among the stages.
Pipelined execution, thanks to the heterogeneous cooperative computation between GPP and CSP, enables to drop per-frame execution time of VGG-16 to 181.2 ms, corresponding to an average throughput of 169.74 GOps/s.

\subsection{ResNet-18}
Very deep neural networks are often difficult to train; the class of Residual Deep Neural Networks aims to solve this issue by providing ``shortcut'' paths between the first and the last layers, improving their correlation at training time, at the cost of a more complex and less regular topology.
ResNet-18 \cite{resnet-18} is one of the first representatives of this class of topologies, which won the 1st place on the ILSVRC 2015 classification task.
These kind of networks are more and more common as they are typically smaller and have lower memory footprint than simpler topologies of equivalent accuracy.


ResNets are built upon a simple basic block consisting in the sum of the results of a chain of several convolutional layers applied on an activation tensor $\mathbf{x}$ with a ``shortcut'' to $\mathbf{x}$ itself, sometimes augmented by a 1$\times$1 convolution layer.
Due to their more complex topology, ResNets are less straightforward to deploy on hardware, however the NeuDNN software stack is able to fully manage this kind of topology. 
%

\begin{figure}[t]
\centering
\includegraphics[width=.65\columnwidth]{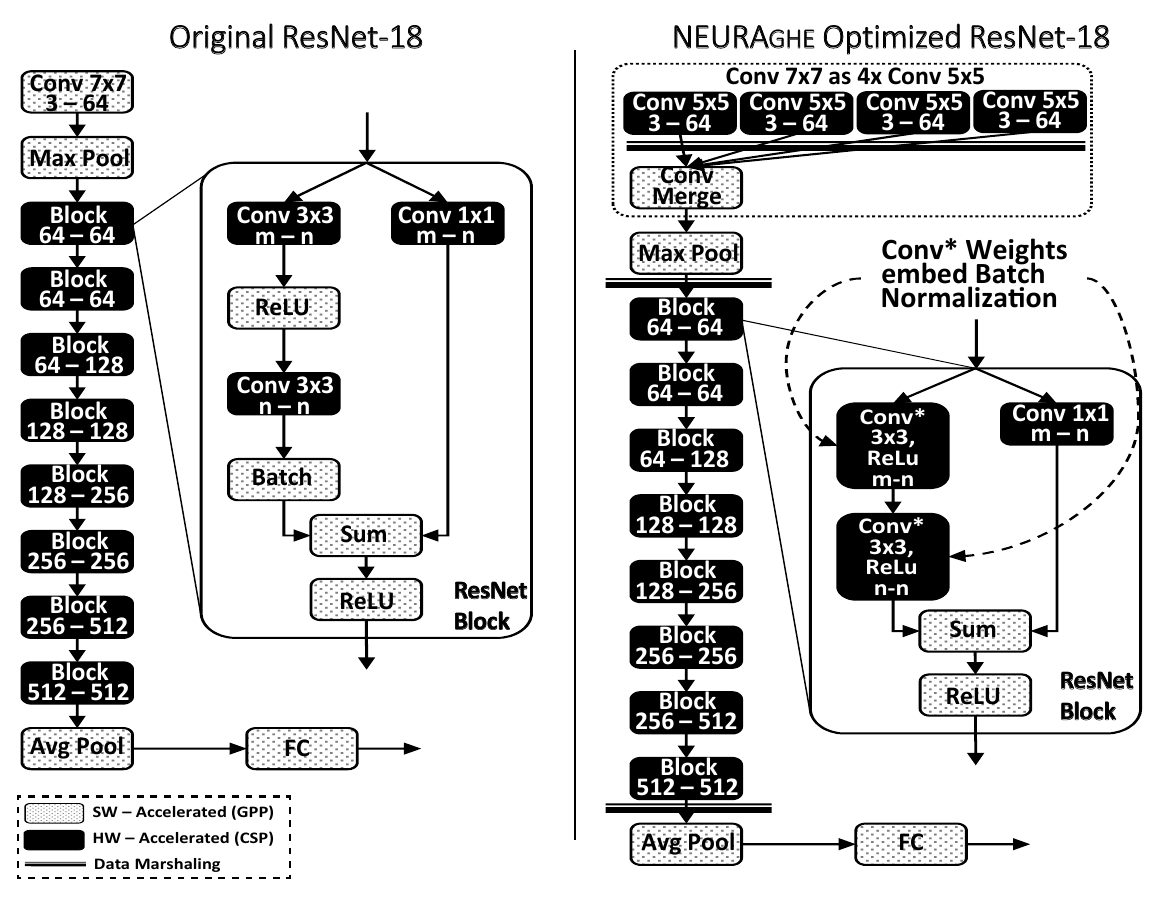}%
\caption{ResNet-18 topologies. Left topology is the original ResNet-18 as proposed by \cite{resnet-18}, while to the right the optimized implementation for \neuraghe{}}%
\label{fig:resnet18}%
\end{figure}

\paragraph{ResNet-18 \neuraghe{} deployment}
Figure \ref{fig:resnet18} shows the ResNet-18 topology.
The left graph shows the original ResNet-18 neural network as proposed by He K.~et~al. \cite{resnet-18} side-by-side with the optimized implementation for Neuraghe.
In black we highlighted the layers that can be forwarded to the Convolution Specific Processor, while the grey boxes are layers that can be executed only on the General Purpose Processor.

In this case, three main modifications were applied to extend the usage of the  Convolution Specific Processor.
First, the 7$\times$7 convolutional layer, which is not natively supported by the Convolutional Engine, was split in four 5$\times$5  convolutional layers followed by a software managed merge layer.
Second, the batch normalization layers, which at inference time simply apply a static pointwise linear operation, where merged with convolution layers by embedding the scaling and translation factors within the convolution weights and biases, respectively \cite{Batch};
ReLU activations are also performed by the  Convolution Engine.
Third, the 1$\times$1 convolutions (not natively supported by the Convolution Engine) are mapped on 3$\times$3 layers.

Similarly to VGG-16, data marshaling layers were added between computations run on CSP and GPP when necessary.
For pointwise operations (e.g. the shortcut merge operations composed of a sum and a ReLu, which runs on the GPP) the interlacing of activations is irrelevant, and thus data marshaling operations around them can be safely skipped.
This is obviously not true for max pooling and fully connected layers.

Like VGG-16, our ResNet-18 implementation use 16-bit fixed point arithmetic for activations, weights, and bias, with Q5.11 format.

\begin{figure}[t]%
\centering
\includegraphics[width=0.9\columnwidth]{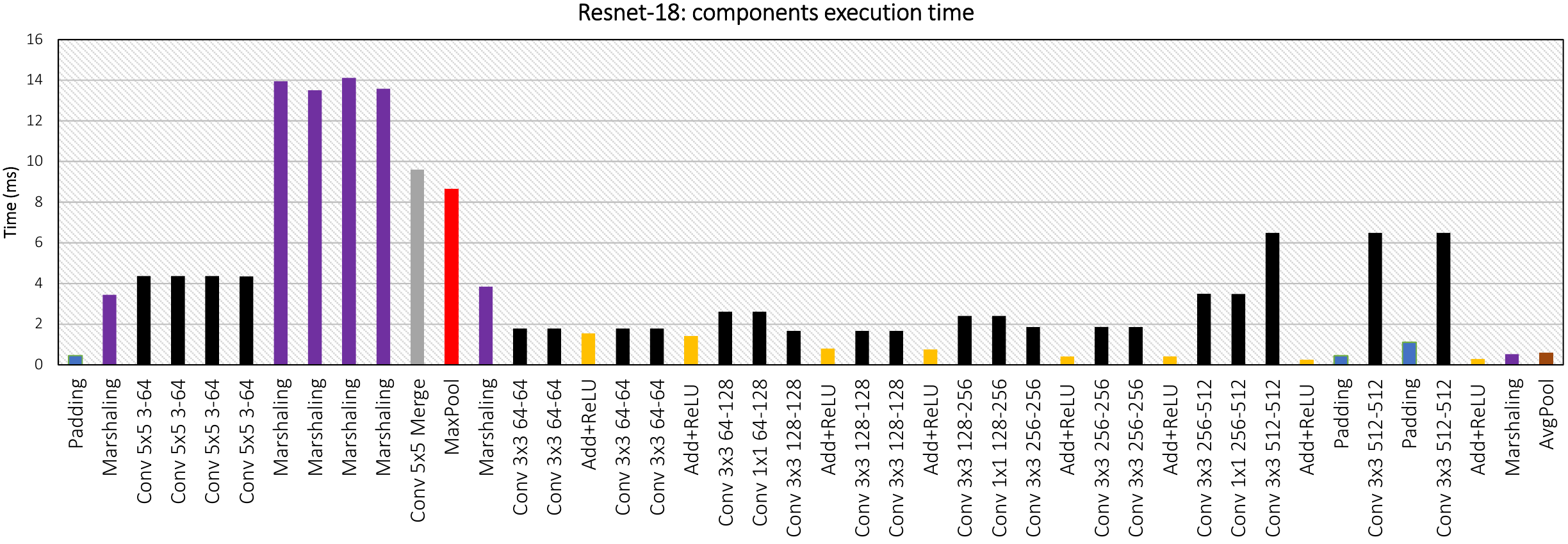}%
\caption{ResNet-18 layer-by-layer profiling.}%
\label{fig:resnet18:profile}%
\end{figure}

\paragraph{ResNet-18 performance analysis}
Figure \ref{fig:resnet18:profile} plots the execution time measured in milliseconds on the \neuraghe{} platform.

The most time-consuming blocks are the four marshaling layers due to the split of the 7$\times$7 convolution in four smaller ones.
Each marshaling action takes up to 14 ms, mainly due to the fact that the amount of data to move and process is significant (64$\times$112$\times$112 pixels).
The second most time consuming layer is the merging of partial results for the emulated 7$\times$7 convolutions, and the max pooling that is in a configuration not supported on the accelerator (3$\times$3 with stride 2).
Both layers take around 9 ms.
5$\times$5 convolutions take $\sim$4 ms, and are penalized by limited number of input activations and the stride 2.
However, thanks to the asynchronous offloading of convolutions to the CSP, these overheads can be partially overlapped with the execution on the accelerator, and can be also parallelized among the two ARM Cortex A9 due to the independence of data marshaling stages with one another.
Thus, while the sum of all the execution time of the layers used to emulate the 7$\times$7 convolution is 92.0 ms, the end-to-end execution time measured is only 51.2 ms, showing up to 40 ms gain due to the cooperative computation of the GPP and the CSP.
The last convolutions are penalized as well, in this case due to the small input feature maps size (only 7$\times$7 pixels) which causes a sub-utilization of the hardware resources in the CE.
Considering the overlaps, the measured end-to-end execution time for the whole neural network inference is 150 ms, equivalent to a frame rate of 6.6 fps.

\begin{figure}[t]
\begin{minipage}[t]{0.45\linewidth}
		\centering
    \includegraphics[width=\linewidth]{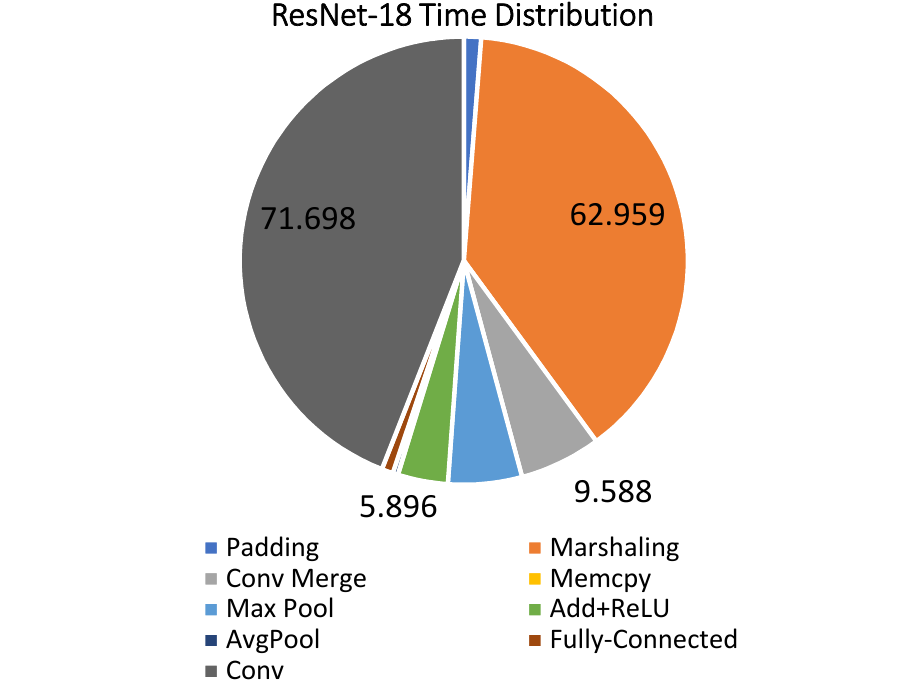}
    \caption{ResNet-18 execution time distribution (milliseconds)}
    \label{fig:resnet18:pipe}
\end{minipage}
\hfill%
\begin{minipage}[t]{0.45\linewidth}
\centering
\includegraphics[width=\linewidth]{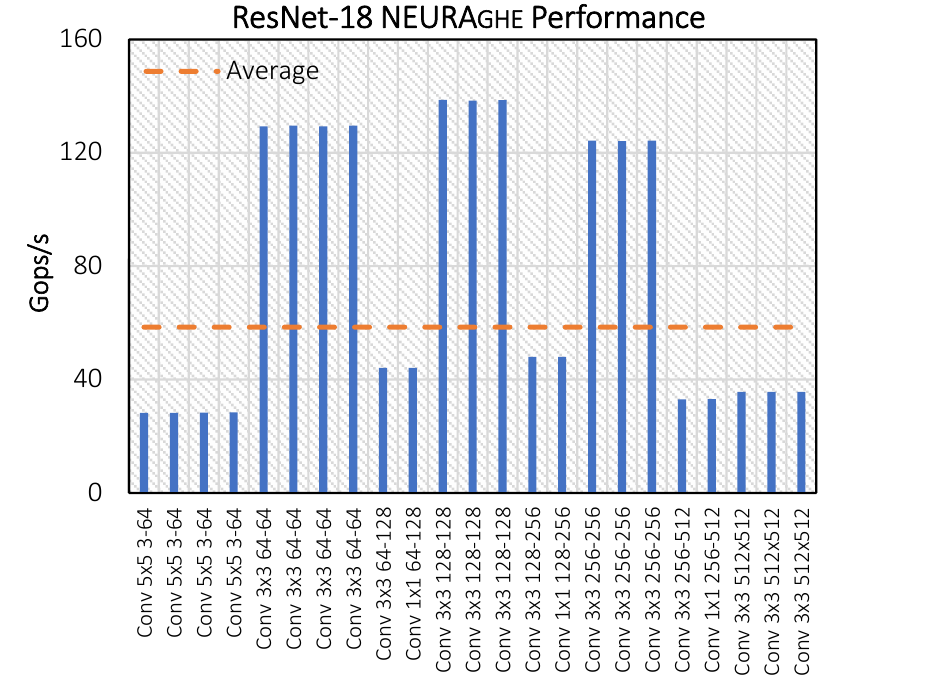}%
\caption{ResNet-18 Convolutional Engine throughput}%
\label{fig:resnet18:gops}%
\end{minipage}
\end{figure}

Figure \ref{fig:resnet18:pipe} shows the time distribution of each component.
The convolutions take around 48\% of the whole time, while 42\% is spent on data-marshaling -- most of it due to the 7$\times$7 convolution.
While the emulated version is not particularly efficient, a pure software execution on the GPP would take up to 176 ms (0.6MOps/s) -- far away from the performance achieved even in a sub-optimal operational region of the CSP.

Finally, Figure \ref{fig:resnet18:gops} shows the measured GOps/s for all the convolutional layers.
For ResNet-18, \neuraghe{} provides up to 140 GOps/s at peak.
On average, throughput drops to 58.4 GOps/s due to two main reason: the striding in output of some  of the convolutions, and the 1$\times$1 convolutions.
This is because in layers with stride values higher than 1, performance is limited by the line buffer functionality. It keeps loading two pixel per cycle from each port but some convolution windows must be discarded, causing idle cycles in the accelerators. 1$\times$1 convolutions are also sub-optimal since a SoP module is under-utilized to perform only 2 MAC operations per cycle, lowering the performance level of the CE.   


\subsection{GPP-accelerated layers performance analysis}
As we discussed, NeuDNN is able not only to exploit the CSP, but also to accelerate other layers that traditionally do not allow optimal mapping on the programmable logic, by means of the capabilities of the ARM Cortex-A9 core.
This is based on two well known methodologies: \textit{thread-level} parallelization, which can be accessed by means of the OpenMP programming model, and \textit{SIMD vectorization}, which is enabled by the NEON vector unit featured by each ARM core, supporting a combined 64- and 128-bit SIMD instruction set for media and signal processing applications.
\begin{figure}[t]
\centering
\includegraphics[width=.75\linewidth]{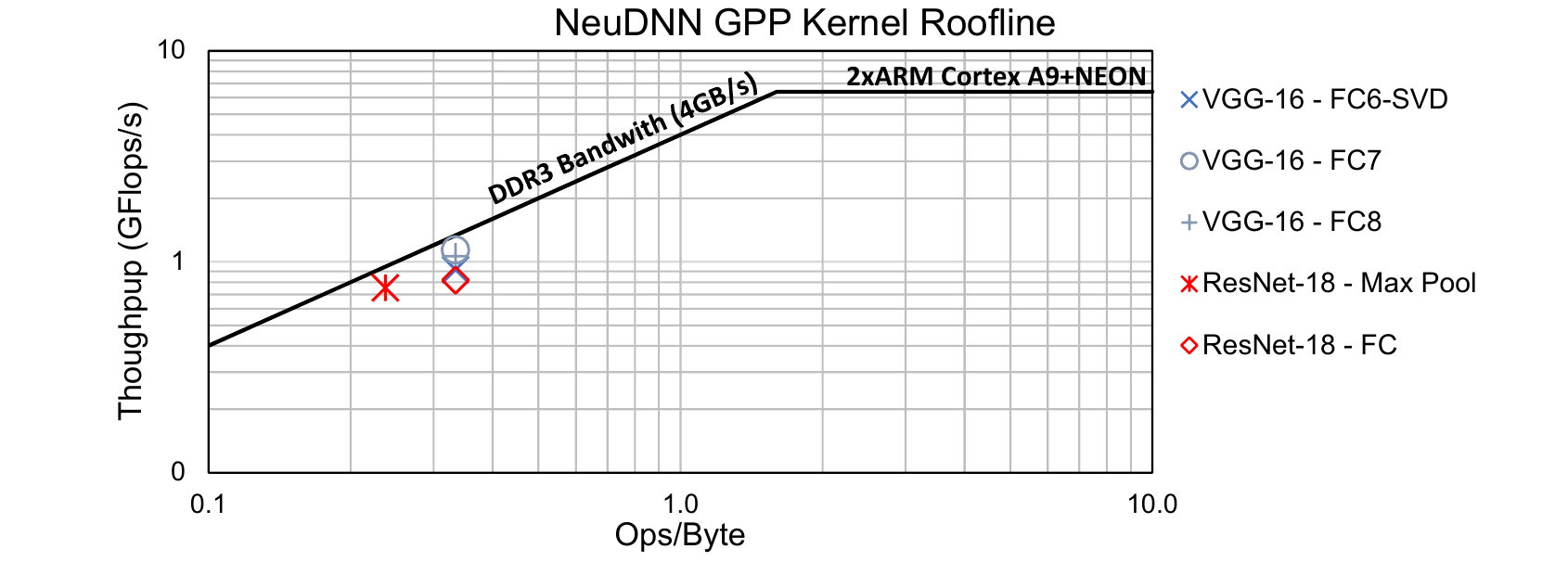}%
\caption{Roofline Model of NeuDNN layers for Xilinx Zynq-7045 SoC}%
\label{fig:roofline}%
\end{figure}

To measure the effectiveness of our implementations, we analyzed the performance generated by the NeuDNN layers executed on the GPP for VGG-16 and ResNet-18 using the well known \textit{roofline model} (Figure \ref{fig:roofline}).
The two ARM Cortex-A9, running at 800MHz, are able to deliver up to 6.4 GFlop/s, and the main memory is limited to 4GB/s.
The \textit{computational density} threshold between \textit{memory-bound} and \textit{compute-bound} operation is in this SoC around 1.5 Op/B.
As recalled in Section \ref{sec:comp_model}, most non-convolutional layers, in particular fully connected layers, are heavily memory bound: each weight is used only once.
This is confirmed in the case of our two target CNNs: we measured a computational density of 0.2-0.3 Op/B, which is well within the memory-bound region.
As can be seen in Figure \ref{fig:roofline}, the NeuDNN software-accelerated implementations are essentially hitting the performance roof set by the memory bandwidth and are therefore optimal given the underlying Zynq architecture.

\subsection{Comparison With State of The Art}
To better understand how the proposed architecture performs with respect to other FPGA accelerators in the state-of-the-art, Table \ref{tab:comparison} provides a comparison with a subset of competitive accelerators dedicated to embedded CNN inference, and deployed on the same Xilinx z-7045 board.
For this reason, all the accelerators show a similar power consumption of 9-10W.
Apart from this, significant differences exist between the various platforms.

In terms of raw performance, \neuraghe{} demonstrates 18-27\% better results than the competing platforms on VGG-16, which is often used as a performance benchmark.
The accelerator proposed by Vernieris~et~al. \cite{LatencyDriven} and Snowflake \cite{Snowflake} claim a performance up to 123 GOps/s and 122 GOps/s, respectively, which is 27\% smaller than the performance of \neuraghe{}, and 18\% smaller than the performance of the accelerator proposed by Qiu~et~al. \cite{GoingDeeper}.
In the case of Vernieris~et~al., performance is mainly limited by the lower operating frequency, which might be attributed to the high-level synthesis methodology, which is not always guaranteed to reach optimal results in terms of implementation.
For what concerns SnowFlake, their operating frequency is the highest, but they use the lowest amount of DSP resources, which negatively impacts their peak performance.
Although they claim that their performance should be scalable by replicating the accelerator design on the same device, a higher occupation of the PL might result in a more congested - and therefore lower frequency - design.
While they report results for ResNet-50, a CNN sharing a similar topology with ResNet-18, it is impossible to perform a direct comparison with their result, as contrarily to the other works they do not report end-to-end performance, but take into account only convolutional layers.
Qiu~et~al. is the strongest competitor to our work, as they deliver up to 138 GOps/s on VGG-16 -- $\sim$18\% less than \neuraghe{}.
The critical advantage provided by our work is that \neuraghe{} fully exploits both the programmable logic and the GPP, ``embracing'' a heterogeneous programming model.
This allows us \textit{i}) to overlap the execution of the fully connected layers and the convolutional layers, and \textit{ii}) to use the NEON extensions on the dual-core ARM Cortex-A9.
\begin{table}[t]
\footnotesize
\begin{threeparttable}
\begin{tabular}{l c c c c}
                   & \multirow{2}{*}{\neuraghe{}}   & \multirow{2}{*}{Qiu~et~al.~\cite{GoingDeeper}} & Gokhale~et~al.  & \multirow{2}{*}{Venieris \& Bouganis~\cite{LatencyDriven}}\\
                   & & & (Snowflake)~\cite{Snowflake} & \\
\toprule
\multirow{2}{*}{\textbf{Platform}}   &Xilinx             &Xilinx            &Xilinx             &Xilinx\\
                 &Zynq Z-7045  &Zynq Z-7045 &Zynq Z-7045  &Zynq Z-7045\\
\textbf{Clock (MHz)} &140MHz  &150MHz         &250MHz         &125MHz\\
\textbf{Power (W)}  &$\sim$10W             &9.63W          &9.61W          &$\sim$10W\\
\toprule
\textbf{DSP}  &864  &780	&256 &900\\
\textbf{LUT}  &100K  &183K	&--- &---\\
\textbf{FF}  &61K	&128K	&--- &---\\
\textbf{BRAM}  &320	&486	&---	&---\\
\toprule
\textbf{Actual Perf.}  &169 (VGG-16) &138 (VGG-16)&122\tnote{1} (ResNet-50)	&123 (VGG-16)\\
\textbf{(GOps/s)} &58 (ResNet-18) &---&120\tnote{1} (AlexNet)&---\\
\multirow{2}{*}{\textbf{Frame/s}}  &  5.5 (VGG-16) & 4.46 (VGG-16)& 17.7\tnote{1} (ResNet-50)	& 4.0 (VGG-16)\\
 &6.6 (ResNet-18)   &---& 100.3\tnote{1} (AlexNet)&---\\
\textbf{End-2-End}  & yes  &yes &no & yes\\
\textbf{Quantization} &16 bit fixed	&16/8/4 bit fixed	&16 bit fixed	&16 bit fixed\\
\end{tabular}
\begin{tablenotes}
	\item[1] Does not include the final fully connected layers.
\end{tablenotes}
\caption{\neuraghe Performance Summary and System Comparison}
\vskip -10pt
\label{tab:comparison}
\end{threeparttable}
\end{table}

%% file: 05_01_hardware_implementation.tex
\subsection{Hardware implementation evaluation}
In the presented configuration, the 16 SoP modules, including 54 DSPs each to produce two output pixels per cycle each, are clocked at 140 MHz. The configuration features four reconfigurable line buffers, each capable of loading up to 128 words (256 pixels). This means that the proposed configuration can process input features which are up to 256 pixel wide. This size is adequate for most of state-of-the art CNN benchmarks. Processing of wider input features requires their partitioning in sub-stripes, by means of dedicated software routines. \par
Table \ref{res_utilization} shows the FPGA resource utilization of the proposed architecture, when mapped on the Zynq XC-Z7045.\par
\begin{table}[ht]
\footnotesize
\begin{minipage}{0.48\textwidth}
\centering
\begin{tabular}{|c |c | c|c|c|c|c|c|c|c|c|}
\hline
    &	DSP  & BRAM  & LUTs  & LUTs         & Regs\\
    &        &       & (logic)   & (SR) & \\
 Used	    &  864   & 320   & 88154   & 11397          & 61250\\
 Avail.	&  900   & 545   & 218600  & 218600         & 437200 \\
 \%&  96.0\% & 58.7\% & 35.1\%   & 16.2\%      & 14.1\%	\\
\hline
\end{tabular}
\caption{Resource utilization on Zynq Z-7045}
\label{res_utilization}
\end{minipage}%
\hfill
\begin{minipage}{0.48\textwidth}
\centering
\begin{tabular}{|c |c | c|c|c|c|}
\hline
    &	DSP  & BRAM  & LUTs  & LUTs         & Regs\\
            &        &       & (logic)   & (SR) & \\
Used	    &  1728   & 640   & 146573   & 22385          & 114261 \\
Avail.	&  2520   & 912   & 274080  & 144000         & 548160 \\
\%	&  68.6\% & 70.2\% & 53.5\%   & 15.6\%      & 20.8\% \\
\hline
\end{tabular}
\caption{Resource utilization on Zynq UltraScale}
\end{minipage}
\end{table}
\vskip -20pt
As may be noticed, the mapping uses 864 out of the 900 DSP blocks available in the device. Thus the proposed configuration uses almost all of the processing power available in the device. BRAM utilization is around 35\%, thus L2 and TCDM size can be comfortably increased if required by the use-case. Also utilization of LUT and registers is low. There is a significant number of LUTs used as shift-registers, due to the internal organization of the line buffer. All the buffer segments that do not need to adapt to different uses and have a static shift path, have been described in HDL to infer use of LUTs, to obtain a  faster and less resource-hungry implementation. It is worth highlighting that the CSP uses only two of the 4 HP ports connecting the programmable logic to the PS and the DDR3. This means that our approach can be scaled easily replicating the number of CSPs in a bigger devices. 
According to our scaling experiments, performed with a Vivado synthesis, a Zynq UltraScale XCZU9EG-2FFVB1156 FPGA would be able to host two CSPs, both clocked at 200 MHz and able to independently access the PS to communicate with the DDR3 memory.

%% file: 06_conclusion.tex
\section{Conclusion}\label{sec:conclusion}
We have presented \neuraghe{}, a Zynq-based processing platform for CNN, specifically designed to improve flexibility and re-usability in different context and for the implementation of different CNN-based algorithms. Our approach relies on the tight interaction between software and hardware. The ARM processing system in the Zynq is not only used for housekeeping tasks, but is also used at its best to perform computation tasks when needed. Moreover, the accelerator implemented in the programmable logic is also controllable via software, integrating a microcontroller in charge of finely managing the basic operations of the other building blocks. We also developed a complete software stack, acting as a distributed runtime on the processing system and on the microcontroller to ease the life of users willing to implement a new CNN case on \neuraghe{}.\par 
We have shown with two different experiments on \neuraghe{} that an approach based on heterogeneous processing, simultaneously exploiting programmable logic and ARM-based processing system, can be used effectively for different reasons. In a first CNN, VGG-16, we have shown that it can be used to improve performance, performing 18\% better than the best competitor in literature. Under the workload imposed by ResNet-18, we have shown that it can be used with success to improve flexibility, implementing on the processing system "irregular" CNN kernels and "adaptation" layers not supported by the accelerator. Our approach is highly-reusable, relying on a completely sw-programmable stack, and scalable, we have successfully implemented two clusters on a Ultrascale+ device, clocked at 200 MHz. Thus, it paves the way for the exploitation of a new acceleration paradigm, relying on hardware-software tight synergy, in the upcoming future of CNN development. It will be a key technique to face challenges posed by next generation of newly appearing CNN algorithms, increasingly irregular and complex, using next-generation of All-Programmable SoCs, increasingly powerful and heterogeneous.